\documentclass{article}
\usepackage{ijcai24}

\usepackage{times}
\usepackage{soul}
\usepackage{url}
\usepackage[hidelinks]{hyperref}
\usepackage[utf8]{inputenc}
\usepackage[small]{caption}
\usepackage{graphicx}
\usepackage{amsmath}
\usepackage{amsthm}
\usepackage{booktabs}
\usepackage{algorithm}
\usepackage{algorithmic}
\usepackage[switch]{lineno}
\usepackage{multirow}
\usepackage{amsmath,amssymb,mathtools}
\usepackage[table]{xcolor}
\usepackage[font=footnotesize]{subfig}

\title{Are Watermarks Bugs for Deepfake Detectors?\\ Rethinking Proactive Forensics}

\author{Anonymous Author(s)}
\author{
	Xiaoshuai Wu
	\and
	Xin Liao\thanks{Corresponding author.}
	\and
	Bo Ou
	\and
	Yuling Liu
	\and
	Zheng Qin
	\affiliations
	College of Computer Science and Electronic Engineering, Hunan University, Changsha, China
	\emails
	\{shinewu, xinliao, oubo, yuling\_liu, zqin\}@hnu.edu.cn
}

\begin{document}

\maketitle

\begin{abstract}
AI-generated content has accelerated the topic of media synthesis, particularly Deepfake, which can manipulate our portraits for positive or malicious purposes. Before releasing these threatening face images, one promising forensics solution is the injection of robust watermarks to track their own provenance. However, we argue that current watermarking models, originally devised for genuine images, may harm the deployed Deepfake detectors when directly applied to forged images, since the watermarks are prone to overlap with the forgery signals used for detection. To bridge this gap, we thus propose AdvMark, on behalf of proactive forensics, to exploit the adversarial vulnerability of passive detectors for good. Specifically, AdvMark serves as a plug-and-play procedure for fine-tuning any robust watermarking into adversarial watermarking, to enhance the forensic detectability of watermarked images; meanwhile, the watermarks can still be extracted for provenance tracking. Extensive experiments demonstrate the effectiveness of the proposed AdvMark, leveraging robust watermarking to fool Deepfake detectors, which can help improve the accuracy of downstream Deepfake detection without tuning the in-the-wild detectors. We believe this work will shed some light on the harmless proactive forensics against Deepfake.
\end{abstract}

\section{Introduction}
\begin{figure}[t]
	\centering
	\includegraphics[width=\linewidth]{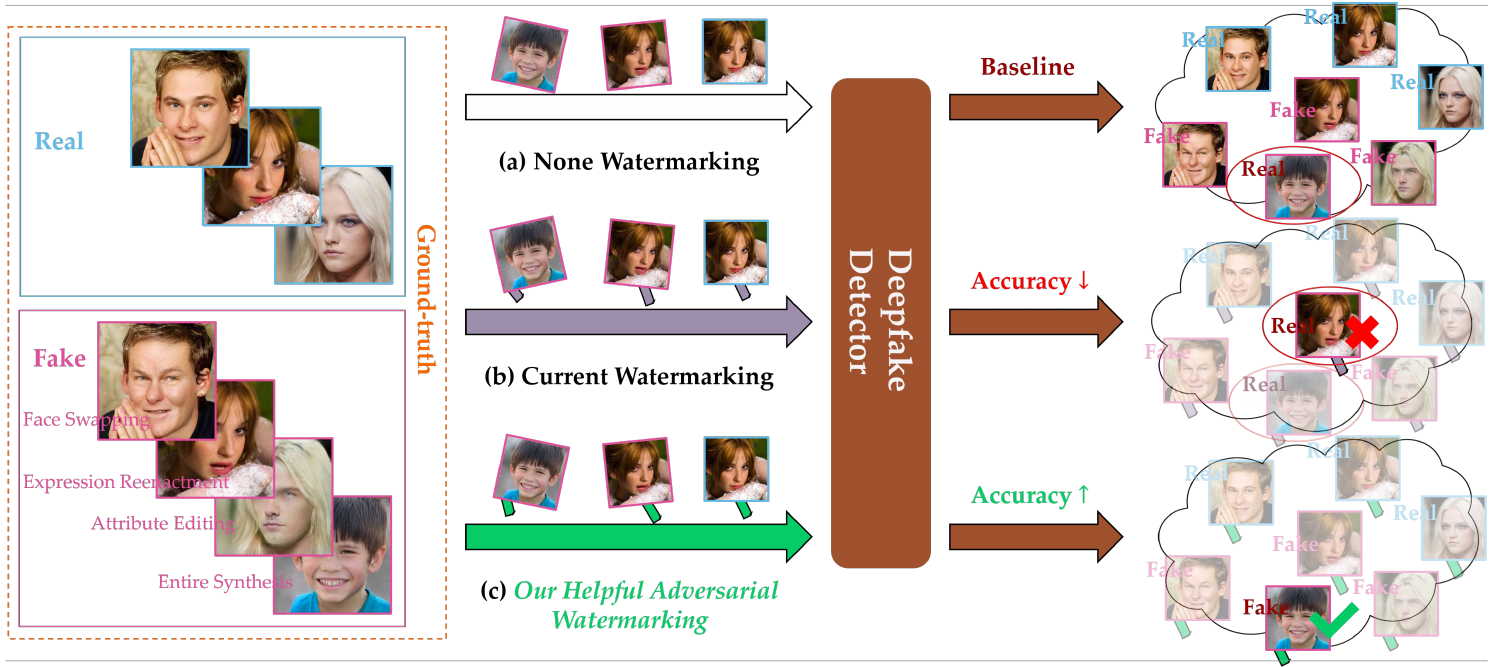}
	\caption{Distinctions between the proposed AdvMark and current watermarking models. (a) Non-watermarked images contributed the baseline of detection performance, where the genuine and forged images may not be easily distinguished. (b) Current watermarking unintentionally degrades the detection performance, since the watermarks are prone to overlap with the forgery signals. (c) Our AdvMark leverages the watermarks to fool Deepfake detectors intentionally, which helps to distinguish between watermarked genuine and forged images without compromising provenance tracking.}
	\label{fig:distinction}
\end{figure}

Large generative models like ChatGPT and Stable Diffusion are transforming the way we communicate, illustrate, and create \cite{hacker2023regulating}. Despite their positive potential in applications such as entertainment, education, art, etc., we cannot neglect the negative impacts. Particularly, the photo-realistic Deepfake \cite{hu2022finfer}, interpreted as ``fake face'' generated through ``deep learning'', has taken false messages and misleading content to a new level, thereby damaging social trust and public interest \cite{juefei2022countering}. In response to this dilemma, emerging standards such as the Coalition for Content Provenance and Authenticity \cite{c2pa} call for embedding provenance information within the released images online \cite{bui2023rosteals}. To build Responsible AI, major tech companies, including Google and Microsoft, are also developing tools to trace the origin and identify synthetically altered or created images \cite{zhao2023generative}. It has never been more important to understand the provenance and authenticity of the images we see, helping us make informed decisions and protect ourselves from deception.

Robust watermarking is the prevalent measure for the task of provenance tracking \cite{zhu2018hidden,tancik2020stegastamp}. In such a watermarking prototype, subtle watermark signals that identify the provenance are imperceptibly embedded into the host image by a watermark encoder, rendering them almost invisible to the human naked eyes. After that, even if the watermarked image has been subjected to severe distortions, the watermark can still be extracted correctly by a watermark decoder in a blind manner, i.e., without knowledge of the original host image. It is also acknowledged that the watermark embedding can be accomplished during the content generation \cite{xiong2023flexible}, similar to the encoder implanted into a specific generative model. It should be noted that in this paper, we focus on more general watermarking methods, which modify the host images and are applicable to arbitrary generative models as well as non-generated images.

In previous studies, proactive watermark injection and passive Deepfake detection were completely independent, without any consideration of their correlation. However, we reveal that in practice, the presence of watermarks leads to more false-negative results when using most passive detectors (see Table~\ref{tab:watermark_effect}), which is indeed contrary to the belief in AI responsibility and does a disservice. This can be explained by the fact that current watermarking models were originally devised for genuine images, where the watermarks are prone to overlap with the forgery signals used for detection. One underlying assumption for the watermark effects is that well-trained detectors are asymmetrically tuned to detect patterns that make an image fake \cite{ojha2023towards}. Consequently, anything else without these patterns would be classified as genuine.

To gain more insights into this intriguing phenomenon, we remark that invisible watermarks are analogous to natural adversarial perturbations \cite{hou2023evading}. Both of them are crafted by subtle signals that are not easily perceptible by human eyes; adversarial perturbations are intended to cause misjudgments of the detectors, while watermarks are designed to enable provenance tracking but unintentionally reduce the detection performance. More fundamentally, from the perspective ``adversarial examples are features'', they can be attributed to non-robust features that are highly predictive for the detectors \cite{ilyas2019adversarial}. Therefore, in another vein, it is possible for the embedded watermarks to be both recoverable and adversarial at the same time.

To this end, we propose AdvMark, which exploits the adversarial vulnerability of passive detectors for good, making the watermarks adversarial to positively improve the performance of Deepfake detectors. The distinctions between the proposed AdvMark and current watermarking models are observable in Figure~\ref{fig:distinction}. Unlike previous harmful adversarial examples and watermarking, our proposed helpful adversarial watermarking deceives the detector into correctly classifying the input image following its ground-truth label. Specifically, in contrast to the original images that were incorrectly predicted, the watermarked images force the detector to report a correct prediction. Meanwhile, the watermarks should not change the detection results of the original correctly predicted images. It is also worth noting that the meaningful watermarks are capable of tracking the image's provenance once extracted, while the meaningless perturbations are not.

Our contributions can be summarized into three-fold:
\begin{itemize}
	\item We present a harmless proactive forensics solution, AdvMark, which leverages robust watermarking to fool Deepfake detectors for the first time, where the embedded watermarks behave both recoverable and adversarial, thereby achieving the purposes of provenance tracking and detectability enhancement simultaneously.
	\item To the best of our knowledge, this is the first attempt to formulate the concept of helpful adversarial watermarking, which amends the watermarked images to improve the accuracy of downstream Deepfake detection without tuning the in-the-wild detectors.
	\item Extensive experiments conducted on well-trained detectors demonstrate the effectiveness of the proposed AdvMark under white- and black-box attacks, on various types of Deepfake, such as face swapping, expression reenactment, attribute editing, and entire synthesis.
\end{itemize}

\section{Related Work}
\subsection{Deepfake Forensics}
Up to now, Deepfake has covered massive types and advanced rapidly, leading to more realistic and convincing face forgeries. To control the dissemination of these, countermeasures are continually developed and can be roughly classified into three categories: passive forensics, proactive defense, and proactive forensics \cite{wu2023sepmark}. Among them, proactive defense performs adversarial attacks to destroy the creation of Deepfake \cite{huang2022cmua}; however, it leaves few forensic clues, as the visually perceptible artifacts are easily perceived and can alert the adversary \cite{chen2021magdr}.

\textbf{Passive Forensics.}\label{sec:detector}
Deep learning-based detectors exploit powerful backbone networks, such as Xception \cite{rossler2019faceforensics++} and EfficientNet \cite{li2021exploring}, and their automatic feature extraction empirically outperforms hand-craft features. CNND \cite{wang2020cnn} incorporates appropriate data augmentations during the network training. FFD \cite{dang2020detection} utilizes informative attention maps that indicate suspicious regions to detect the forgery. PatchForensics \cite{chai2020makes} develops the patch-based detector with limited receptive fields to exaggerate local artifacts in patch regions. MultiAtt \cite{zhao2021multi} captures local discriminative features from multiple attentive regions to identify subtle and local differences between genuine and forged images. RFM \cite{wang2021representative} erases several sensitive facial regions to guide the detector to allocate more attention to the representative regions. RECCE \cite{cao2022end} mines common compact representations of genuine faces from the reconstruction perspective. SBI \cite{shiohara2022detecting} follows the self-supervised paradigm where the more general forgeries can be produced on-the-fly.

\textbf{Proactive Forensics.}
To combat Deepfake proactively, FakeTagger \cite{wang2021faketagger} is derived from robust watermarking, where the embedded watermarks can survive both before and after Deepfake, enabling the tracking of the provenance of watermarked images. FaceSigns \cite{neekhara2022facesigns} introduces a semi-fragile watermarking framework, and its proactive detection is realized through verifying the extracted watermarks. SepMark \cite{wu2023sepmark} enables the embedded watermark to be extracted at different levels of robustness, achieving the purposes of provenance tracking and proactive detection simultaneously. Likewise, BiFPro \cite{liu2023bifpro} extends watermarking into diverse forensic scenarios where the watermark manifests either fragility or robustness. The embedded watermarks can be identity-related messages \cite{zhao2023proactive,wang2023robust,guan2023building}. Other than deep watermarking, Source-ID-Tracker \cite{lin2022source} employs steganography to hide the original images within their forged versions. PIMD \cite{asnani2022proactive} and MaLP \cite{asnani2023malp} add the learned templates to conduct proactive detection and location, respectively. Moreover, AFP \cite{yu2021artificial} and Faketracer \cite{sun2022faketracer} poison the training data, resulting in watermarked generative models.

\subsection{Adversarial Attacks}
Seminal work \cite{szegedy2013intriguing} discloses the vulnerable nature of deep neural networks to adversarial examples. Transferability is an intriguing property of adversarial examples; deliberately perturbed inputs generated for the surrogate model can mislead other victim models' predictions. Adversarial attacks can be categorized as white-box and black-box, depending on whether the target model can be accessed (e.g., architecture and parameters). Moreover, untargeted attacks aim to cause the model to misclassify inputs regardless of the specific category. Accordingly, targeted attacks aim to cause the model to misclassify inputs into another specific category. Notably, the generative attacks \cite{poursaeed2018generative,xiao2018generating} leverage a trainable generator to directly generate desired perturbations, which are ingeniously similar to current watermarking models and require only a single forward pass in the inference stage, as opposed to instance-specific iterative optimization attacks. For more details on this avenue, the review \cite{zhao2023revisiting} is recommended.

\section{Motivation \& Insight}
\textbf{Motivation.} The previous proactive defense method \cite{wang2022deepfake} takes the downstream task of passive forensics into consideration, where the attacked unrealistic Deepfake should be effortlessly recognized by the detectors. It is noteworthy that the proactive defense methods \cite{zhu2023information,zhang2024dual} also consider proactive forensics, providing the adversarial perturbations with the forensic clue that is capable of provenance tracking. However, the research regarding the correlation between proactive forensics and passive forensics remains substantially unfilled. To study how the watermarks impact Deepfake detectors, we report the detection results of the images watermarked by MBRS \cite{jia2021mbrs}, FaceSigns, and SepMark, which represent robust, semi-fragile, and multipurpose watermarking, respectively. Table~\ref{tab:watermark_effect} shows that the detectors have a clear tendency to predict the watermarked forged images as genuine. To rectify this, a harmless proactive forensics solution is urgently needed, which should concurrently take the downstream Deepfake detectors into consideration, where the watermarked images should be more easily distinguished by the Deepfake detectors, relative to non-watermarked images.

\textbf{Insight.} Although the term ``adversarial'' leaves an indelible impression of its malicious functionality, adversarial training for benign purposes can effectively boost model robustness. What's more, along with the belief in ``adversarial for good'' \cite{al2024adversarial}, if harnessed in the right manner, the adversarial nature can be utilized to improve image recognition models \cite{xie2020adversarial}, amend the accuracy level of neural models \cite{yu2023adversarial}, boost object detection performance \cite{sun2022rethinking}, protect watermarks against removal \cite{liu2022watermark}, prevent automatic painting imitation \cite{liang2023adversarial} and Deepfake generation \cite{wang2022anti}. In a similar vein, we delve into robust watermarking to exploit the adversarial vulnerability of Deepfake detectors for good. Note that this paper is not focusing on developing a new robust watermarking model or a new Deepfake detector. Instead, we propose a plug-and-play targeted attack procedure for fine-tuning any robust watermarking to help improve the accuracy of downstream Deepfake detection without tuning the detectors (see Figure~\ref{fig:definition}). Concretely, during our adversarial fine-tuning procedure, only the parameters of watermarking models are updated, and Deepfake detectors keep unchanged. Fine-tuning robust watermarking into adversarial watermarking involves several challenges and key factors. First, it's crucial that the detector attack exhibits efficacy in both white- and black-box scenarios, with adversarial transferability occupying a pivotal position. Secondly, the watermark extraction should still be conducted successfully to enable provenance tracking. Lastly, the visual quality of the watermarked images should remain pleasing.

\begin{table}[t]
	\centering
	\tiny
	\begin{tabular}{llccc}
		\toprule
		& & Xception & EfficientNet & CNND \\ \hline \hline
		\multirow{2}{*}{Clean}      & Real ACC & 98.19 & 99.11 & 99.96 \\
		& Fake ACC & 20.82 & 52.73 & 27.87 \\ \hline
		\multirow{2}{*}{JPEG}      & Real ACC & 84.49 & 1.13 & 99.89 \\
		& Fake ACC & 32.15 & 99.89 & 18.06 \\ \hline
		\multirow{2}{*}{Gaussian Noise}      & Real ACC & 99.93 & 25.07 & 100.0 \\
		& Fake ACC & 0.00 & 82.61 & 0.32 \\ \hline
		\multirow{2}{*}{\begin{tabular}{@{}l@{}}MBRS\\ \cite{jia2021mbrs}\end{tabular}} & Real ACC & 98.73 & 98.83 & 99.96 \\
		& Fake ACC & 17.53 & 45.29 & 23.41 \\ \hline
		\multirow{2}{*}{\begin{tabular}{@{}l@{}}FaceSigns\\ \cite{neekhara2022facesigns}\end{tabular}} & Real ACC & 98.62 & 98.83 & 99.82 \\
		& Fake ACC & 10.91 & 38.60 & 23.09 \\ \hline
		\multirow{2}{*}{\begin{tabular}{@{}l@{}}SepMark\\ \cite{wu2023sepmark}\end{tabular}} & Real ACC & 98.44 & 99.47 & 99.96 \\
		& Fake ACC & 15.30 & 28.26 & 24.11 \\ \hline
	\end{tabular}
	\caption{Accuracy of well-trained Deepfake detectors tested on non-watermarked/JPEG-compressed/Gaussian noisy/watermarked images in detecting separate real and fake subsets. We have the following observations: \romannumeral1) the detectors predict almost all the genuine images with high accuracy, regardless of whether they are watermarked or not, but struggle with diverse types of forged images (e.g., SimSwap, FOMM, StarGAN, and StyleGAN); \romannumeral2) the watermarks distort the images in a counter-intuitive fashion, which is actually different from JPEG compression and Gaussian noise; and \romannumeral3) the watermarked forged images are more likely to be predicted as real by the detectors, compared to the non-watermarked forged counterparts. The experimental setup here is consistent with Section~\ref{sec:experiment}.}
	\label{tab:watermark_effect}
\end{table}

\section{Methodology}
\subsection{Problem Formulation} 
\textbf{Adversarial Perturbations Harm Detectors.} Let $x$ denote the clean host image, and $y$ denote its ground-truth label. We use $\hat x$ to represent the adversarial image, i.e., $\hat x = x+\eta$, where $\eta$ is the imperceptible perturbation with the $L_p$-norm constraint $\lVert\eta \rVert_p\le \epsilon, \; p\in\{0,2,\infty\}$. Suppose there is a perturbation generator $\mathcal{P}(\cdot)$ generating the perturbations directly, i.e., $\hat x = x+\mathcal{P}(x)$, and a Deepfake detector $\mathcal{D}(\cdot)$ predicting the label of the input. We recall the vanilla adversarial attack, where the objective is to mislead the detector into making an incorrect prediction by 
\begin{equation}
	\max_\mathcal{P} \mathbb{E}_{(x,y)} \; \mathcal{F}(\mathcal{D}(x+\mathcal{P}(x)), y),
\end{equation}
where $\mathcal{F}$ denotes the commonly used cross-entropy loss or binary cross-entropy loss for the task of Deepfake detection. It should be noted that the perturbations generated by $\mathcal{P}(\cdot)$ are typically constrained by a small $L_\infty$-norm budget, to maintain the lower bound of visual quality but may not align well with human perception, with respect to perturbed images.

\begin{figure}[t]
	\centering
	\includegraphics[width=0.9\linewidth]{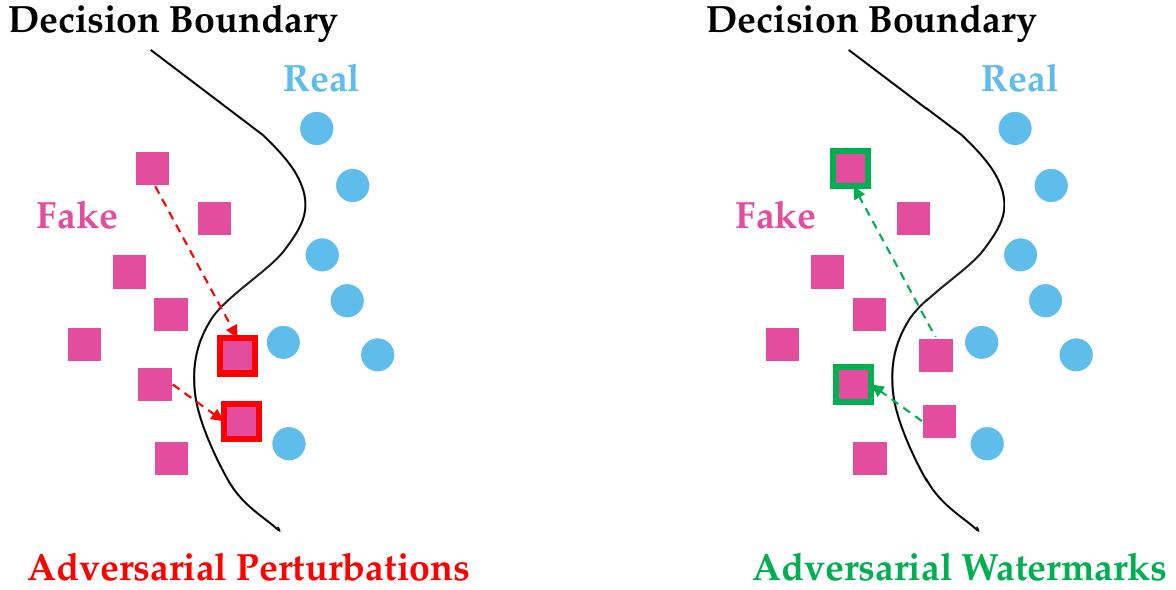}
	\caption{The sketch of adversarial perturbations vs. adversarial watermarks. Left: The objective of adversarial perturbations is to make correctly predicted inputs result in wrong detection outcomes. Right: The objective of adversarial watermarks is to make original incorrectly predicted inputs yield correct detection outcomes.}
	\label{fig:definition}
\end{figure}

\textbf{Adversarial Watermarks Help Detectors.}
Given a watermark encoder $En(\cdot)$ with the input of the host image $x$ and watermark $w$, we can obtain the watermarked image $x_w$, i.e., $x_w=En(x, w)$. A corresponding watermark decoder $De(\cdot)$ should be able to extract the watermark $w$ exactly from the image $x_w$. Nevertheless, considering that the watermarked image is sometimes delivered and distributed through the dirty channel, it is proposed to introduce a noise layer $\mathcal{N}(\cdot)$ to distort the image $x_w$ with diverse data augmentations, where the distorted image $\widetilde{x_w}=\mathcal{N}(x_w)$. In this way, the robust decoder can extract an approximate watermark from the image $\widetilde{x_w}$, i.e., $\widetilde{w}=De(\widetilde{x_w})$, meaning $\widetilde{w}\approx w$. Formally, the objective of our adversarial watermarking, which helps the detector classify watermarked images, can be formulated as
\begin{equation}
	\begin{split}
		\min_{En, De} \mathbb{E}_{(x,y,w)} \; &[\mathcal{F}(\mathcal{D}(En(x, w)), y) + \mathcal{G}(En(x, w), x) \\&+ \mathcal{H}(De(\mathcal{N}(En(x, w))), w)],
	\end{split}
\end{equation}
where $\mathcal{G}$ denotes the loss function used for image similarity measures (e.g., mean-squared error loss and LPIPS loss \cite{zhang2018unreasonable}), and $\mathcal{H}$ represents the mean-squared error loss or binary cross-entropy loss, which usually characterizes the watermark extraction as either a regression or a classification problem, respectively.

\subsection{AdvMark Framework} 
\begin{figure}[t]
	\centering
	\includegraphics[width=\linewidth]{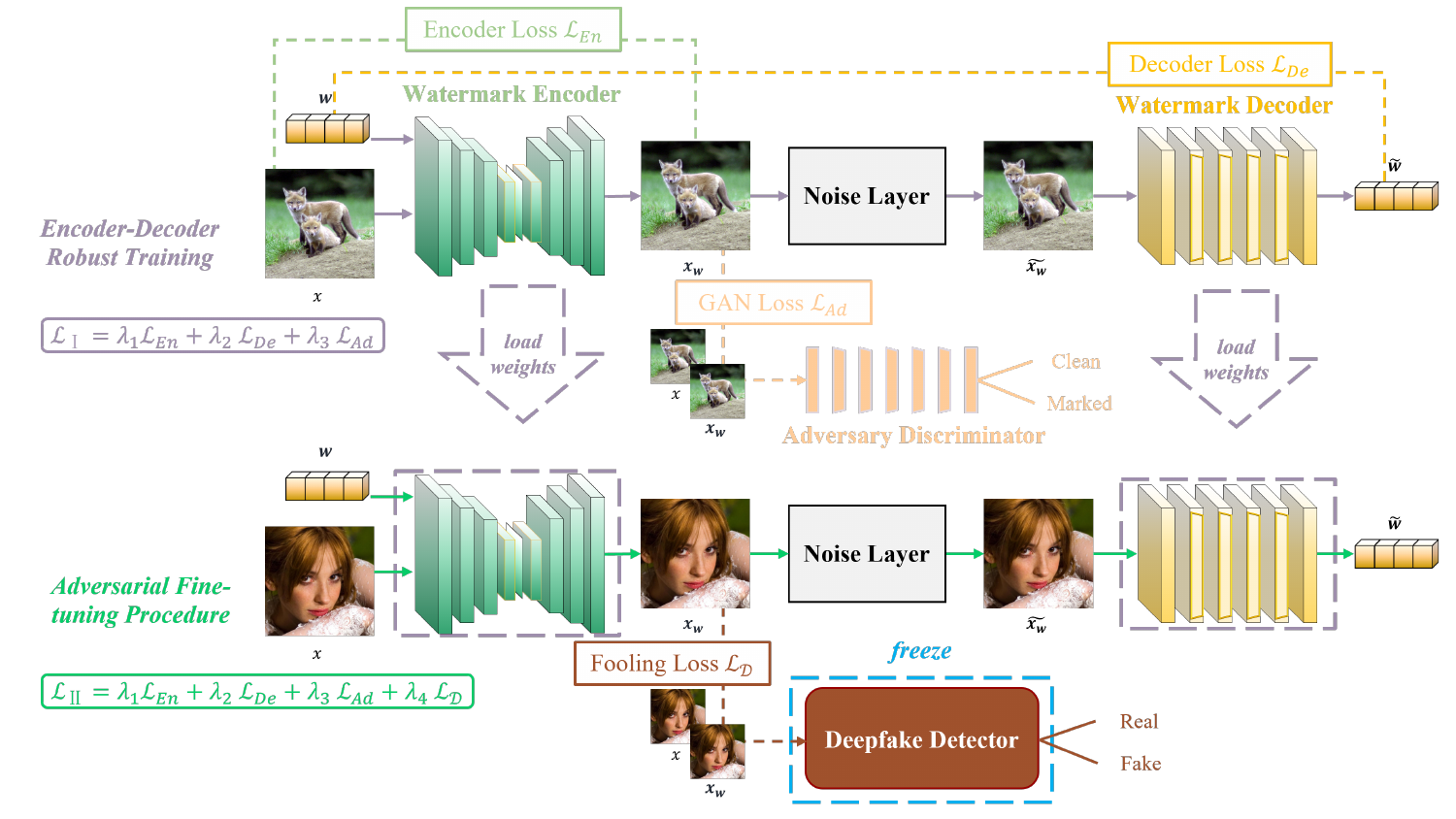}
	\caption{Overview of our proposed AdvMark. The training process consists of two stages. In Stage \uppercase\expandafter{\romannumeral1}, the encoder and decoder are jointly trained end-to-end, obtaining the pre-trained encoder and decoder that serve as robust watermarking. In Stage \uppercase\expandafter{\romannumeral2}, the encoder and decoder are fine-tuned end-to-end by fooling a surrogate Deepfake detector, aiming at transforming robust watermarking into adversarial watermarking. During inference, only the final watermark encoder and decoder will be adopted.}
	\label{fig:AdvMark}
\end{figure}

As illustrated in Figure~\ref{fig:AdvMark}, the pipeline of our proposed AdvMark consists of five components: \emph{Watermark Encoder}, \emph{Noise Layer}, \emph{Watermark Decoder}, \emph{Adversary Discriminator} and downstream \emph{Deepfake Detector}. Specifically, we showcase the functionalities of each component as follows:
\begin{enumerate}
	\item[1)] The encoder $En$ receives a batch of the host image $x\in \mathbb{R}^{B\times 3\times H\times W}$ (which belongs to either genuine or forged, labeled by $y\in \{0, 1\}^{B\times 1}$), and random watermark bits $w\in \{0, 1\}^{B\times L}$ as input. It produces the watermarked image $x_{w}$, which has intriguing adversarial properties for helping the downstream Deepfake detectors.
	\item[2)] The noise layer $\mathcal{N}$ augments the watermarked image $x_{w}$ with various simulated or real distortions, resulting in the distorted image $\widetilde{x_w}$. It is parameterless but differentiable, which connects the encoder $En$ and decoder $De$ to facilitate end-to-end robust training.
	\item[3)] The decoder $De$ extracts the watermark $\widetilde{w}$ from the distorted image $\widetilde{x_w}$, where $\widetilde{w}\approx w$, to reliably track the provenance of watermarked image $x_w$. Note that it also participates in updating the encoder $En$ to generate the image that is more conducive to watermark extraction.
	\item[4)] The adversary discriminator $Ad$ attempts to distinguish between the host image $x$ and the watermarked image $x_w$. Conversely, the encoder $En$ should mislead the discriminator so that $Ad$ cannot distinguish $x_w$ from $x$, to acquire better visual quality of the watermarked image.
	\item[5)] The surrogate detector $\mathcal{D}$ classifies the watermarked image $x_w$ as genuine or forged. In the case of $\mathcal{D}(x)\neq y$, the watermarked image generated by the encoder $En$ makes sense when $\mathcal{D}(x_w)=y$, reflecting the enhanced forensic detectability. This watermarked image is even easier to be detected by other downstream detectors.
\end{enumerate}

Below we will provide a detailed description of the pre-training stage of robust watermarking, the fine-tuning stage of adversarial watermarking, and the inference stage of final watermark encoder and decoder in turn.

\textbf{Encoder-Decoder Robust Training.} For the watermark encoder $En$, its watermarked image $x_w$ should be visually similar to the original host image $x$: 
\begin{equation}
	\mathcal{L}_{En}=L_{2}(x,x_w)=L_{2}(x,En(\theta;x,w)),
\end{equation}
where $\theta$ denotes the trainable parameters of the encoder.

For the watermark decoder $De$, its extracted watermark $\widetilde{w}$ should be identical to the original embedded watermark $w$, from the distorted image $\widetilde{x_w}$ augmented by noise layer $\mathcal{N}$:
\begin{equation}
	\mathcal{L}_{De}=L_{2}(w,\widetilde{w})=L_{2}(w,De(\phi;\widetilde{x_w})),
\end{equation}
where 
\begin{equation}
	\widetilde{x_w}=\mathcal{N}(x_w)=\mathcal{N}(En(\theta;x,w)),
\end{equation}
and $\phi$ represents the trainable parameters of the decoder.

For the adversary discriminator $Ad$, there are several loss variants for stabilizing the GAN training \cite{jolicoeur2018relativistic}, and for ease of presentation, here we describe the standard GAN loss \cite{goodfellow2014generative}:
\begin{equation}
	\mathcal{L}_{Adv}=\text{log}(1-Ad(\gamma;En(x,w)))+\text{log}(Ad(\gamma;x)),
\end{equation}
where $\gamma$ indicates the trainable parameters of the discriminator. Note that the discriminator is trained alternately with the encoder-decoder, where the encoder $En$ is also updated by
\begin{equation}
	\mathcal{L}_{Ad}=\text{log}(Ad(x_w))=\text{log}(Ad(En(\theta;x,w))).
\end{equation}

To sum up, the total loss for Stage \uppercase\expandafter{\romannumeral1} can be formulated by
\begin{equation}
	\mathcal{L}_{\uppercase\expandafter{\romannumeral1}}=\lambda_{1}\mathcal{L}_{En}+\lambda_{2}\mathcal{L}_{De}+\lambda_{3}\mathcal{L}_{Ad},
\end{equation}
where $\lambda_{1}$, $\lambda_{2}$, $\lambda_{3}$ are the weights for respective loss terms. After training, we get the pre-trained encoder $En_{\uppercase\expandafter{\romannumeral1}}$ and decoder $De_{\uppercase\expandafter{\romannumeral1}}$. We can also freely adopt the released models\footnote{\scriptsize\url{https://github.com/jzyustc/MBRS}}\textsuperscript{,}\footnote{\scriptsize\url{https://github.com/sh1newu/SepMark}} as the pre-trained encoder and decoder for this stage. More implementation details can be found in Section~\ref{sec:implementation}.

\textbf{Adversarial Fine-tuning Procedure.} The advent of highly realistic generated content can easily deceive human eyes. Manually examining a large volume of images through observation is extremely expensive, even when the defects are obvious. Therefore, it is preferable to develop automatic Deepfake detectors on the machine end. However, with the proliferation of generated images, merely verifying their authenticity seems insufficient without understanding their provenance, since it's difficult to differentiate between images created for malicious intents and those created for neutral or positive purposes. Building upon the watermarking at Stage \uppercase\expandafter{\romannumeral1}, a responsible individual or social network can embed the provenance evidence into the published images using his/her watermark encoder $En_{\uppercase\expandafter{\romannumeral1}}$. Unfortunately, the proactive injection will harm the passive detectors, which are more prevalent and widely deployed in the wild than the solely watermark decoder $De_{\uppercase\expandafter{\romannumeral1}}$. This is indeed contrary to the belief in AI responsibility and does a disservice. In light of the above, we propose a plug-and-play targeted attack procedure that fine-tunes robust watermarking into adversarial watermarking, aiming to help improve the accuracy of downstream Deepfake detection by deliberately fooling the passive detectors (i.e., crossing the decision boundary; see Figure~\ref{fig:definition}).

To achieve the fooling objective, we freeze a surrogate detector $\mathcal{D}$ and force the detector to predict the correct label $y$ with respect to the watermarked image $x_w$ by updating $En_{\uppercase\expandafter{\romannumeral1}}$:
\begin{equation}
	\mathcal{L}_{\mathcal{D}}=\mathcal{F}(\mathcal{D}(x_w),y)=\mathcal{F}(\mathcal{D}(En_{\uppercase\expandafter{\romannumeral1}}(\theta;x,w)),y),\label{eq:9}
\end{equation}
where $\mathcal{F}$ denotes the loss function (e.g., binary cross-entropy loss) used to train the original detector $\mathcal{D}$.

To summarize, the total loss for Stage \uppercase\expandafter{\romannumeral2} is formulated by
\begin{equation}
	\mathcal{L}_{\uppercase\expandafter{\romannumeral2}}=\lambda_{1}\mathcal{L}_{En}+\lambda_{2}\mathcal{L}_{De}+\lambda_{3}\mathcal{L}_{Ad}+\lambda_{4}\mathcal{L}_{\mathcal{D}},
\end{equation}
where $\lambda_{4}$ is the weight for the fooling loss term. After fine-tuning, we get the final encoder $En_{\uppercase\expandafter{\romannumeral2}}$ and decoder $De_{\uppercase\expandafter{\romannumeral2}}$. Detecting the watermarked images amended by $En_{\uppercase\expandafter{\romannumeral2}}$ will help improve detection accuracy without tuning the detectors.

\textbf{Model Inference.} In the inference process, only the well-trained watermark encoder $En_{\uppercase\expandafter{\romannumeral2}}$ and decoder $De_{\uppercase\expandafter{\romannumeral2}}$ are adopted. Consequently, depending on our adversarial watermarks, we can categorize the harmless proactive forensics into the following cases.

Case 1: If $\widetilde{w}\not\approx w$, the suspicious image is either not watermarked by the encoder $En_{\uppercase\expandafter{\romannumeral2}}$ at all or has been extremely altered by significant distortions, reflecting that it is impossible to track its provenance by the decoder $De_{\uppercase\expandafter{\romannumeral2}}$.

Case 2: If $\widetilde{w}\approx w$, the image has been watermarked by the encoder $En_{\uppercase\expandafter{\romannumeral2}}$, and we can extract the watermark $\widetilde{w}$ to serve as evidence of the image's provenance, and effortlessly determine whether it is genuine or forged by the detector $\mathcal{D}$, which we refer to as enhanced forensic detectability.

Case 3: Even in a more challenging setting where the detector $\mathcal{D}^{\prime}$ was unseen during the training process, the watermarked image $x_w$ is also more easily distinguished by $\mathcal{D}^{\prime}$, which is known as adversarial transferability.
 
\begin{table*}[t]
	\centering
	\tiny
	\setlength{\tabcolsep}{3.5pt}
	\begin{tabular}{lcccccccccc}
		\toprule
		& Xception & EfficientNet & CNND & FFD & PatchForensics & MultiAtt & RFM & RECCE & SBI & Average\\ \hline \hline
		Clean      & 98.19/20.82 & 99.11/52.73 & 99.96/27.87 & 97.17/59.67 & 99.40/33.43 & 85.23/20.36 & 39.41/82.33 & 54.36/60.38 & 87.57/30.06 & 84.49/43.07 \\
		JPEG      & 84.49/32.15 & 1.13/99.89 & 99.89/18.06 & 99.96/0.11 & 100.0/0.00 & 91.25/13.39 & 98.87/0.21 & 92.92/15.16 & 90.93/17.17 & 84.38/21.79 \\
		Gaussian Noise      & 99.93/0.00 & 25.07/82.61 & 100.0/0.32 & 96.53/5.45 & 100.0/0.00 & 77.58/20.54 & 85.98/5.67 & 29.99/86.54 & 99.96/0.11 & 79.45/22.36 \\ \hline
		MBRS      & 98.73/17.53 & 98.83/45.29 & 99.96/23.41 & 94.33/54.21 & 99.36/30.38 & 84.77/20.01 & 42.92/79.78 & 69.19/44.97 & 89.34/26.17 & 86.38/37.97 \\ 
		+ Vanilla Fine-tuning      & 98.41/18.41 & 97.77/52.83 & 99.93/19.12 & 97.06/41.04 & 99.65/31.94 & 86.23/19.37 & 43.73/79.14 & 63.99/49.50 & 87.89/27.90 & 86.07/37.69 \\ 
		+ AdvMark (Xception)      & \cellcolor{gray!20}99.82/99.82$^*$ & 99.04/95.04 & 99.96/5.67 & 96.49/1.88 & 98.26/69.26 & 87.68/0.92 & 78.15/24.68 & 74.79/82.75 & 95.08/5.88 & 92.14/42.88 \\
		+ AdvMark (Efficient.)      & 99.08/15.58 & \cellcolor{gray!20}100.0/99.89$^*$ & 99.96/21.00 & 98.37/22.06 & 84.77/54.64 & 88.07/18.56 & 63.99/77.87 & 79.04/38.56 & 92.81/26.13 & 89.57/41.59 \\ 
		+ AdvMark (CNND)      & 98.97/15.90 & 41.57/51.13 & \cellcolor{gray!20}99.68/99.47$^*$ & 97.59/49.86 & 86.97/83.14 & 89.09/25.28 & 96.14/84.24 & 86.69/9.67 & 94.05/10.30 & 87.86/47.67 \\
		+ AdvMark (FFD)      & 99.65/10.34 & 76.06/30.84 & 99.93/13.63 & \cellcolor{gray!20}86.19/94.19$^*$ & 47.27/82.37 & 83.04/23.23 & 65.93/75.99 & 71.53/33.60 & 85.91/34.70 & 79.50/44.32 \\
		+ AdvMark (PatchFor.)      & 98.94/8.53 & 94.79/84.45 & 99.96/7.19 & 98.41/24.04 & \cellcolor{gray!20}100.0/99.96$^*$ & 86.61/13.88 & 99.72/25.32 & 77.83/30.67 & 89.73/18.63 & 94.00/34.74 \\
		+ AdvMark (MultiAtt)      & 99.15/35.87 & 96.74/94.83 & 99.96/13.03 & 98.48/40.51 & 98.12/64.45 & \cellcolor{gray!20}100.0/100.0$^*$ & 91.68/94.44 & 99.65/80.63 & 99.01/25.53 & 98.09/61.03 \\ 
		+ AdvMark (RFM)      & 99.08/15.30 & 92.25/72.10 & 99.96/23.90 & 96.42/48.90 & 96.67/60.62 & 88.00/18.02 & \cellcolor{gray!20}100.0/100.0$^*$ & 70.61/42.07 & 91.43/25.99 & 92.71/45.21 \\
		+ AdvMark (RECCE)      & 99.72/18.20 & 35.30/94.55 & 99.96/19.09 & 99.79/34.60 & 100.0/57.01 & 95.75/19.26 & 52.83/70.93 & \cellcolor{gray!20}100.0/100.0$^*$ & 98.55/11.01 & 86.88/47.18 \\
		+ AdvMark (SBI)      & 99.58/41.75 & 99.26/89.31 & 99.96/20.40 & 99.01/22.45 & 96.74/34.60 & 96.18/46.57 & 90.79/80.88 & 95.89/88.17 & \cellcolor{gray!20}100.0/99.43$^*$ & 97.49/58.17 \\
		+ AdvMark (Ensemble)      & \cellcolor{gray!20}99.89/99.08$^*$ & 88.99/96.42 & 99.96/21.21 & 83.32/28.58 & 99.82/68.87 & \cellcolor{gray!20}99.96/99.72$^*$ & \cellcolor{gray!20}100.0/99.93$^*$ & 91.78/98.51 & 98.58/63.88 & 95.81/75.13 \\
		\hline
		SepMark      & 98.44/15.30 & 99.47/28.26 & 99.96/24.11 & 94.62/59.45 & 99.15/37.50 & 87.75/18.02 & 36.30/81.06 & 59.81/51.70 & 89.13/22.45 & 84.96/37.54 \\ 
		+ Vanilla Fine-tuning      & 98.09/19.51 & 98.37/44.16 & 99.96/25.11 & 92.71/61.05 & 98.34/39.87 & 86.54/17.46 & 34.81/82.65 & 62.54/50.46 & 86.37/26.35 & 84.19/40.74 \\  
		+ AdvMark (Xception)      & \cellcolor{gray!20}99.82/99.68$^*$ & 98.76/70.08 & 99.96/29.39 & 92.42/66.43 & 99.40/43.66 & 89.73/64.48 & 29.53/82.54 & 73.02/95.04 & 93.41/87.46 & 86.23/70.97 \\
		+ AdvMark (Efficient.)      & 98.55/23.09 & \cellcolor{gray!20}99.96/99.82$^*$ & 99.96/26.81 & 92.81/62.61 & 99.15/36.05 & 87.75/24.08 & 36.83/81.94 & 68.31/66.47 & 89.59/40.47 & 85.88/51.26 \\ 
		+ AdvMark (CNND)      & 99.29/21.10 & 98.94/52.62 & \cellcolor{gray!20}99.89/97.84$^*$ & 90.01/62.64 & 97.73/47.42 & 85.30/16.82 & 38.53/83.00 & 59.42/57.47 & 93.09/31.83 & 84.69/52.30 \\
		+ AdvMark (FFD)      & 98.76/17.99 & 99.33/45.89 & 99.96/25.92 & \cellcolor{gray!20}98.90/95.64$^*$ & 99.68/39.24 & 86.30/20.33 & 37.71/83.82 & 62.71/58.11 & 92.28/25.81 & 86.18/45.86 \\
		+ AdvMark (PatchFor.)      & 98.69/18.70 & 99.22/37.96 & 99.96/23.94 & 96.00/86.93 & \cellcolor{gray!20}99.96/99.82$^*$ & 88.31/16.11 & 38.63/88.81 & 69.02/46.03 & 91.82/22.38 & 86.85/48.96 \\
		+ AdvMark (MultiAtt)      & 99.33/46.81 & 97.70/83.82 & 99.89/23.80 & 92.92/63.07 & 95.72/54.43 & \cellcolor{gray!20}99.86/99.93$^*$ & 33.68/86.69 & 95.68/97.42 & 95.36/70.29 & 90.02/69.58 \\
		+ AdvMark (RFM)      & 98.26/18.52 & 99.11/56.34 & 99.93/22.66 & 94.44/68.56 & 99.65/43.31 & 86.23/21.85 & \cellcolor{gray!20}99.15/100.0$^*$ & 59.42/57.22 & 85.38/30.21 & 91.29/46.52 \\
		+ AdvMark (RECCE)      & 99.43/27.37 & 98.76/60.30 & 99.96/27.02 & 92.35/67.28 & 98.37/39.52 & 97.63/55.13 & 39.02/83.32 & \cellcolor{gray!20}99.96/99.96$^*$ & 93.84/50.11 & 91.04/56.67 \\
		+ AdvMark (SBI)      & 98.83/28.36 & 99.01/56.41 & 99.96/24.47 & 89.34/56.34 & 98.94/37.36 & 90.86/32.51 & 34.07/83.11 & 71.78/75.53 & \cellcolor{gray!20}99.58/98.83$^*$ & 86.93/54.77 \\
		+ AdvMark (Ensemble)      & \cellcolor{gray!20}99.40/97.91$^*$ & 98.87/87.36 & 99.96/30.49 & 93.94/75.18 & 99.65/40.08 & \cellcolor{gray!20}99.22/99.47$^*$ & \cellcolor{gray!20}99.04/99.86$^*$ & 92.56/98.12 & 93.20/85.87 & 97.32/79.37 \\
		\hline
	\end{tabular}
	\caption{Detection accuracy (Real/Fake ACC $\uparrow$) under white \& black-box attacks. ``\colorbox{gray!20}{$^*$}'' indicates seen detector during training.} 
	\label{tab:detector_attack}
\end{table*}
\begin{table*}[t]
	\centering
	\tiny
	\setlength{\tabcolsep}{3.5pt}
	\begin{tabular}{lccccccccccccccc}
		\toprule
		& Identity & JPEG & Resize & GB & MB & Brightness & Contrast & Saturation & Hue & Dropout & SP & GN & Average & PSNR (dB) & SSIM \\ \hline \hline
		MBRS      & 0.0001 & 0.253    & 1.764 & 3.528 & 2.290 & 1.002 & 1.143 & 0.001 & 0.001 & 0.048 & 31.87 & 13.10 & 4.583 & 44.31 & 0.972 \\
		+ Vanilla Fine-tuning      & 0.000 & 0.160    & 2.615 & 0.204 & 0.196 & 1.035 & 0.984 & 0.001 & 0.001 & 0.037 & 37.34 & 13.06 & 4.636 & 43.98 & 0.975\\ 
		+ AdvMark (Xception)      & 0.000 & 0.909    & 0.058 & 3.246 & 7.115 & 1.021 & 0.672 & 0.001 & 0.280 & 0.153 & 9.293 & 6.736 & 2.457 & 38.72 & 0.909 \\
		+ AdvMark (Efficient.)      & 0.000 & 0.689    & 0.004 & 0.538 & 0.370 & 1.100 & 0.868 & 0.001 & 0.031 & 0.013 & 14.26 & 13.30 & 2.598 & 41.88 & 0.955 \\
		+ AdvMark (CNND)      & 0.0001 & 1.469    & 0.358 & 14.67 & 19.21 & 0.772 & 0.227 & 0.001 & 0.425 & 0.246 & 15.48 & 2.251 & 4.592 & 34.00 & 0.848 \\
		+ AdvMark (FFD)      & 0.001 & 0.349    & 0.782 & 1.643 & 2.675 & 0.847 & 0.489 & 0.001 & 0.002 & 0.029 & 21.48 & 0.362 & 2.388 & 34.67 & 0.894 \\
		+ AdvMark (PatchFor.)      & 0.000 & 0.455    & 4.961 & 5.675 & 13.43 & 0.998 & 0.465 & 0.000 & 0.000 & 0.007 & 25.74 & 1.262 & 4.416 & 36.70 & 0.912 \\
		+ AdvMark (MultiAtt)      & 0.000 & 0.169    & 0.058 & 0.652 & 0.711 & 1.043 & 0.865 & 0.000 & 0.040 & 0.007 & 16.44 & 5.533 & 2.127 & 36.20 & 0.876 \\
		+ AdvMark (RFM)      & 0.000 & 0.743    & 0.699 & 1.728 & 1.813 & 1.065 & 0.958 & 0.001 & 0.002 & 0.049 & 28.70 & 9.678 & 3.786 & 43.14 & 0.967 \\
		+ AdvMark (RECCE)      & 0.000 & 1.586    & 0.011 & 0.188 & 0.151 & 1.121 & 1.197 & 0.003 & 0.231 & 0.014 & 8.891 & 15.94 & 2.444 & 39.10 & 0.908 \\
		+ AdvMark (SBI)      & 0.000 & 0.974    & 0.025 & 0.561 & 0.395 & 1.169 & 0.984 & 0.001 & 0.156 & 0.051 & 12.26 & 12.47 & 2.421 & 38.84 & 0.926 \\ 
		+ AdvMark (Ensemble)      & 0.000 & 1.830   & 1.640 & 3.632 & 3.850 & 1.014 & 1.003 & 0.002 & 1.473 & 0.019 & 19.48 & 7.792 & 3.478 & 37.38 & 0.916 \\ \hline
		SepMark      & 0.005 & 0.006  & 0.0001  & 0.0001 & 0.001 & 0.011 & 0.004 & 0.006 & 0.008 & 0.112 & 0.0004 & 0.066 & 0.018 & 38.52 & 0.930\\ 
		+ Vanilla Fine-tuning      & 0.000 & 0.004    & 0.000 & 0.000 & 0.000 & 0.003 & 0.000 & 0.000 & 0.000 & 0.000 & 0.000 & 0.088 & 0.008 & 40.66 & 0.948 \\  
		+ AdvMark (Xception)      & 0.000 & 0.027   & 0.0001 & 0.000 & 0.000 & 0.001 & 0.000 & 0.000 & 0.000 & 0.0001 & 0.0001 & 0.111 & 0.012 & 37.97 & 0.924\\
		+ AdvMark (Efficient.)      & 0.000 & 0.274    & 0.001 & 0.0003 & 0.0001 & 0.006 & 0.001 & 0.000 & 0.000 & 0.003 & 0.005 & 1.128 & 0.118 & 42.13 & 0.961 \\
		+ AdvMark (CNND)      & 0.000 & 0.576    & 0.003 & 0.001 & 0.0004 & 0.005 & 0.001 & 0.000 & 0.000 & 0.001 & 0.001 & 1.341 & 0.161 & 39.63 & 0.945 \\
		+ AdvMark (FFD)      & 0.000 & 1.891    & 0.016 & 0.007 & 0.011 & 0.010 & 0.004 & 0.000 & 0.0001 & 0.011 & 0.014 & 3.432 & 0.450 & 44.49 & 0.977 \\
		+ AdvMark (PatchFor.)      & 0.000 & 0.119    & 0.001 & 0.000 & 0.000 & 0.003 & 0.001 & 0.000 & 0.000 & 0.001 & 0.001 & 0.572 & 0.058 & 41.57 & 0.955 \\ 
		+ AdvMark (MultiAtt)      & 0.000 & 0.007   & 0.000 & 0.000 & 0.000 & 0.002 & 0.000 & 0.000 & 0.000 & 0.0001 & 0.0001 & 0.112 & 0.010 & 38.46 & 0.925\\
		+ AdvMark (RFM)      & 0.000 & 0.288    & 0.001 & 0.0003 & 0.0001 & 0.003 & 0.0004 & 0.000 & 0.000 & 0.000 & 0.0003 & 1.008 & 0.108 & 42.12 & 0.961 \\
		+ AdvMark (RECCE)      & 0.000 & 0.122    & 0.0003 & 0.000 & 0.000 & 0.002 & 0.000 & 0.000 & 0.000 & 0.001 & 0.0003 & 0.551 & 0.056 & 39.87 & 0.947 \\
		+ AdvMark (SBI)      & 0.000 & 0.156   & 0.001 & 0.000 & 0.000 & 0.004 & 0.001 & 0.000 & 0.000 & 0.0004 & 0.001 & 0.700 & 0.072 & 40.12 & 0.949 \\
		+ AdvMark (Ensemble)      & 0.000 & 0.054   & 0.0001 & 0.0001 & 0.000 & 0.003 & 0.000 & 0.000 & 0.000 & 0.0001 & 0.0001 & 0.287 & 0.029 & 38.18 & 0.928 \\ \hline
	\end{tabular}
	\caption{Extraction error rate (BER $\downarrow$) \& PSNR ($\uparrow$) and SSIM ($\uparrow$) of watermarked images. ``$0.000$'' indicates error-free extraction. Gaussian Blur, Median Blur, Salt Pepper, and Gaussian Noise are abbreviated as GB, MB, SP, and GN, respectively.} 
	\label{tab:BER_PSNR}
\end{table*}

\section{Experiments}\label{sec:experiment}
\subsection{Dataset Preparation} 
We collect the real faces sourced from CelebA-HQ \cite{karras2017progressive} and resize them to the resolution of $256\times 256$. These real faces are then manipulated by recent Deepfake generative models: SimSwap \cite{chen2020simswap} for face swapping, FOMM \cite{siarohin2019first} for expression reenactment, and StarGAN \cite{choi2018stargan} for attribute editing. We further utilize the entire synthesized faces provided by StyleGAN \cite{karras2019style}, which are also resized to $256\times 256$. These four fake categories are evenly sampled to balance the numbers of real and fake faces. More specifically, there are equal numbers of faces in real and fake subsets, and we divide them into training, validation, and testing, respectively, referencing the official split $24183/2993/2824$. Due to the page limit, more details and experiments can be found in the supplement at \url{https://github.com/sh1newu/AdvMark}.

\subsection{Implementation Details.}\label{sec:implementation} As the proposed work is not focusing on developing a new robust watermarking model, we directly adopt MBRS and the robust branch of SepMark as the pre-trained encoder and decoder in Stage \uppercase\expandafter{\romannumeral1}, and also as our baselines. To show that the performance gain is due to our adversarial fine-tuning procedure rather than training for more iterations, we establish additional baselines where the two watermarking models undergo a vanilla fine-tuning procedure (without fooling loss) on the dataset. Moreover, we utilize nine well-trained Deepfake detectors, as described in Section~\ref{sec:detector}, to validate the capability of forensic detectability and adversarial transferability. For the hyper-parameter settings of MBRS and SepMark, we strictly follow their original implementations. For example, besides the difference in the length of the watermark bits ($256$-bit for MBRS and $128$-bit for SepMark), the noise layers they use are also inconsistent. To be specific, MBRS includes \emph{Identity}, \emph{JPEG}, and \emph{simulated JPEG}, while SepMark contains all the noises listed in Table~\ref{tab:BER_PSNR}. Therefore, we will compare our AdvMark with the respective baselines for each to mitigate the model bias introduced by different backbones. Lastly, the whole fine-tuning lasted for $10$ epochs with a batch size of $8$, and we set the weight of the fooling loss $\lambda_{4}$ to $0.1$.

\subsection{Detector Attack}
The benign functionality of adversarial watermarking is confirmed in both white- and black-box scenarios. The evaluation metric here is the accuracy (ACC) of the detector $\mathcal{D}$:
\begin{equation}
	ACC(\mathcal{D})=\frac{1}{B} \sum_{i=1}^{B} |\mathcal{D}(x^{i\times 3\times H\times W})-y^{i\times 1}| \times 100\%,
\end{equation}
where 
\begin{equation}
	\mathcal{D}(x)=\left\{
	\begin{array}{ll}
		0 & \text{   if   } x \text{   is predicted as real},\\
		1 & \text{   if   } x \text{   is predicted as fake},
	\end{array}
	\right.
\end{equation}
and $y\in \{0, 1\}^{B\times 1}$ means the ground-truth label, $|\cdot|$ denotes the absolute value. For a comprehensive evaluation, we report the Real/Fake ACC on the separate real and fake subsets. The higher the value, the better the detection performance.

\textbf{White-box Attack.}
In the white-box scenario, we evaluate the ``training accuracy" with the detector seen during training. When the watermark encoder-decoder and Deepfake detector are in the same camp, i.e., the detector is already known by the watermarking developer, the surrogate detector used can be consistent with the downstream detector. By observing Table~\ref{tab:detector_attack}, we note that the white-box detector without any tuning can get significant performance gain in detecting the images watermarked by AdvMark. The baselines MBRS and SepMark are similar to the image processing operations, e.g., JPEG compression and Gaussian noise, which are harmful to the detector, where they distort the host image in different fashions. The fine-tuned baselines also have inferior detection performance compared to the clean host images without watermarking. By contrast, our helpful AdvMark turns the watermark from an enemy into a friend, ensuring nearly $100\%$ accuracy for the detector in the white-box setting. This is also the case when several detectors, e.g., Xception, MultiAtt, and RFM, are included for the adversarial fine-tuning.

\textbf{Black-box Attack.}
We further evaluate the ``testing accuracy" with the detector unseen during training, in the black-box scenario. Table~\ref{tab:detector_attack} shows that compared to the clean host images, the watermarked images using AdvMark are more easily distinguished by other unseen detectors, thanks to the transferability of adversarial watermarks. For example, using MultiAtt or SBI as the surrogate detector, the resulting detection performance with AdvMark is higher than that without watermarking, on average. We notice that AdvMark is inferior to the baselines on a few unseen detectors, which may be due to the different decision logic shared by the detectors. E.g., most detectors mainly identify distinct fake patterns in the images, while the exception RECCE seeks to discriminate what is real. Since AdvMark performs quite well in the white-box setting, we speculate that it has the potential to overfit to the seen detector, but on some unseen detectors it may have ruined the detection. Suggested by ensemble attacks \cite{poursaeed2018generative}, we can attack an ensemble of detectors, which is more conducive in the real world. By simultaneously attacking Xception, MultiAtt, and RFM, in the black-box setting we obtain an overall improved detection performance, which is indeed better than attacking only one detector.

\begin{figure}[t!]
	\centering
	\subfloat[\parbox{15mm}{Backbone: MBRS}]{\includegraphics[width=0.32\columnwidth]{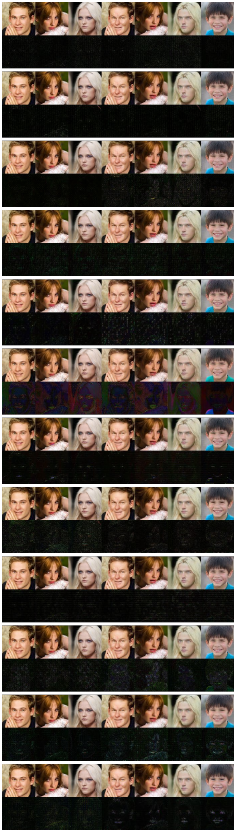}} \hspace{10mm}
	\subfloat[\parbox{15mm}{Backbone: SepMark}]{\includegraphics[width=0.32\columnwidth]{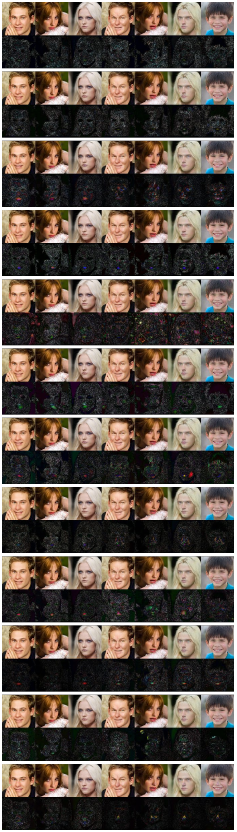}}
	\caption{Visualizations on watermarked images and normalized residuals. From top to bottom, refer to the default order in Table~\ref{tab:BER_PSNR}.}
	\label{fig:visual_quality}
\end{figure}

\subsection{Watermark Extraction \& Visual Quality}
Since our AdvMark focuses on improving the forensic detectability of watermarked images without compromising provenance tracking, we expect as few performance degradations in Bit Error Rate (BER) as possible. Suppose that the embedded watermark $w\in \{0, 1\}^{B\times L}$ and the extracted watermark $\widetilde{w}\in \{0, 1\}^{B\times L}$, formally,
\begin{equation}
	BER(w,\widetilde{w})=\frac{1}{B}\times \frac{1}{L}\times \sum_{i=1}^{B} \sum_{j=1}^{L} |w^{i\times j}-\widetilde{w}^{i\times j}|\times 100\%.
\end{equation}
A lower BER indicates stronger robustness in watermark extraction. From Table~\ref{tab:BER_PSNR}, we can see that the BER of AdvMark is comparable to the baselines, and the performance gaps are almost negligible, regardless of the surrogate detector used. Interestingly, the adversarially fine-tuned MBRS enhances the robustness against unseen distortions, e.g., salt pepper and Gaussian noise, which is likely due to the redundancy of embedded adversarial watermarks. The robustness resists against distortion attacks may depend on the utilized watermarking backbone; it shall be even greater when using a more advanced watermarking backbone.

Although AdvMark accomplishes both provenance tracking and detectability enhancement, we have to admit that these benefits come at the cost of visual quality to a certain extent. As seen in Table~\ref{tab:BER_PSNR}, the watermarked images using AdvMark exhibit distinct PSNR and SSIM values that can vary for different surrogate detectors. Despite these trade-offs, the subjective visual quality should still be pleasurable. Figure~\ref{fig:visual_quality} visualizes the watermarked images and their min-max normalized residual signals relative to clean host images. By observation, we speculate that the rationale behind the watermark residuals may be attributed to the real or fake patterns associated with the decision logic of the detector. While the watermarked images have distinct residual patterns, the subjective visual quality still aligns well with human perception.

\section{Conclusion}
In this work, we make the first attempt to bridge proactive forensics and passive forensics, the two previously uncorrelated studies, and propose the concept of helpful adversarial watermarking, reflecting the belief in ``adversarial for good''. Without training the watermarking model from scratch, AdvMark can serve as a plug-and-play procedure, seamlessly integrating with existing watermarking. By fine-tuning robust watermarking into adversarial watermarking, we do not harm the performance of passive detectors while fulfilling proactive forensics, achieving harmless provenance tracking and concurrently enhancing forensic detectability. We also investigate the transferability of adversarial watermarks, allowing AdvMark to generalize the watermarking to unseen Deepfake detectors, further validating its effectiveness in the wild. In our future work, we plan to conduct more analyses of the watermark effects on other downstream tasks, such as tamper localization \cite{zhang2024editguard}.


\section*{Acknowledgments}
This work is supported by National Natural Science Foundation of China (Grant Nos. U22A2030, U20A20174), National Key R\&D Program of China (Grant No. 2022YFB3103500), Hunan Provincial Funds for Distinguished Young Scholars (Grant No. 2024JJ2025).

\bibliographystyle{named}
\bibliography{ijcai24}

\end{document}